\documentclass{article}

\usepackage[final]{tccml_neurips_2020}

\usepackage[utf8]{inputenc} 
\usepackage[T1]{fontenc}    
\usepackage{hyperref}       
\usepackage{url}            
\usepackage{booktabs}       
\usepackage{amsfonts}       
\usepackage{nicefrac}       
\usepackage{microtype}      
\usepackage{graphicx}

\title{Characterization of Industrial Smoke Plumes from Remote Sensing Data}

\author{%
  Michael Mommert\thanks{\texttt{michael.mommert@unisg.ch}}\\
  University of St.\ Gallen\\
  \And
  Mario Sigel\\
  Sociovestix Labs Ltd.\\
  \And 
  Marcel Neuhausler\\
  ISS Inc.\\
  \And
  Linus Scheibenreif\\
  University of St.\ Gallen \\
  \And
  Damian Borth\\
  University of St.\ Gallen\\
}

\begin{document}

\maketitle

\begin{abstract}
The major driver of global warming has been identified as the anthropogenic release of greenhouse gas (GHG) emissions from industrial activities. The quantitative monitoring of these emissions is mandatory to fully understand their effect on the Earth's climate and to enforce emission regulations on a large scale.
In this work, we investigate the possibility to detect and quantify industrial smoke plumes from globally and freely available multi-band image data from ESA's Sentinel-2 satellites. Using a modified ResNet-50, we can detect smoke plumes of different sizes with an accuracy of 94.3\%. The model correctly ignores natural clouds and focuses on those imaging channels that are related to the spectral absorption from aerosols and water vapor, enabling the localization of smoke. 
We exploit this localization ability and train a U-Net segmentation model on a labeled sub-sample of our data, resulting in an Intersection-over-Union (IoU) metric of 0.608 and an overall accuracy for the detection of any smoke plume of 94.0\%; on average, our model can reproduce the area covered by smoke in an image to within 5.6\%. The performance of our model is mostly limited by occasional confusion with surface objects, the inability to identify semi-transparent smoke, and human limitations to properly identify smoke based on RGB-only images.
Nevertheless, our results enable us to reliably detect and qualitatively estimate the level of smoke activity in order to monitor activity in industrial plants across the globe. 
Our data set and code base are publicly available.
\end{abstract}

%
%
\section{Introduction}
\label{sec:introduction}

Global warming poses a major threat to the social and economic stability of human civilization. 
Paradoxically, most of the currently observed climate changes are mainly driven by human activities. The unrestricted consumption of fossil fuels leads to steadily rising levels of atmospheric CO$_2$ and other greenhouse gases (GHG), resulting in the subsequent trapping of heat in Earth's atmosphere and water bodies \citep[see the][for a summary]{IPCCAR5}. The reduction of GHG emissions is mandatory to limit long-term damage to Earth's climate.

Being able to globally monitor GHG emissions would allow us to (1) obtain a deeper understanding of its effects on the climate and (2) to enforce environmental protection emission quotas and emission trading schemes. However, the direct measurement of the amount of GHG emissions is extremely expensive, especially on large scales. In addition, legal requirements to report industrial emissions vary significantly. A method to globally quantify industrial emissions would improve our picture of GHG emissions, enable their systematic monitoring, and
inform policy makers.

This work investigates the possibility to quantify GHG  emissions by using industrial smoke plumes based on satellite imagery as a proxy. The goal of this ongoing project is to establish a pipeline to monitor the state and level of activity of industrial plants using readily available remote sensing data in an effort to estimate their GHG emissions in combination with environmental data.

Our contribution\footnote{The code base for this work is available at \href{https://github.com/HSG-AIML/IndustrialSmokePlumeDetection}{github.com/HSG-AIML/IndustrialSmokePlumeDetection}; the complete data set is available at \href{http://doi.org/10.5281/zenodo.4250706}{zenodo.org}.}
with regard to climate change monitoring is threefold: (1) we compile a large scale annotated data set of active industrial sites with additional segmentation masks for a subset of these smoke plumes, (2) we present a modified ResNet-50 approach able to detect active smoke plumes with an accuracy of 94.3\% and finally, (3) we utilize a U-Net approach to segment smoke plumes and measure their areal projections on average within 5.6\% of human manual annotations.

%
%
\section{Related Work}
\label{sec:related_work}

The detection of smoke from remote sensing imaging data in the past has mostly been restricted
to the identification of wild fires \citep[see][for a review]{Szapowski2019}. A common method for the 
detection of wildfire smoke uses multi-thresholding of multi-spectral imaging data \citep[e.g.,][]{Randriambelo1998},
exploiting the spectral characteristics of smoke plumes by hand-crafting corresponding features.
With the rise of machine learning, supervised and unsupervised learning methods have been implemented
to automatically detect wildfires and their smoke plumes \citep{Jain2020}.
Other efforts to characterize smoke plumes
use physical dispersion or statistical models \citep[see][for a review]{Jerrett2005}, or they utilize ground-based
observations from consumer grade camera systems 
in combination with algorithmic solutions \citep{Hsu2018} 
or deep learning approaches \citep{Hohberg2015, Hsu2020, Jiao2020}. While 
cheap and easy to implement, such camera systems are 
severely limited in the wavelength range they can observe in and have to be 
deployed for each site individually. Multi-spectral observations from space can monitor large areas and 
make use of additional spectral information.

This work focuses on the detection and characterization of smoke
plumes from industrial activities based on Sentinel-2 MultiSpectral Instrument 
remote sensing data \citep{MSIuserguide} in an end-to-end deep learning approach. 
The advantage of our approach
is the combination of multi-spectral remote sensing data that are available freely and on a global scale 
with a flexible learning approach that not only detects smoke, but is able to quantify the 
amount of activity using a segmentation model. In contrast to traditional multi-thresholding methods, our deep-learning approach can be easily adopted to other data sets (other continents, other satellites) without the need for labor-intensive hand-crafting of the spectral features of smoke and the background.

\begin{figure}[t]
  \centering
  \includegraphics[height=1.55in]{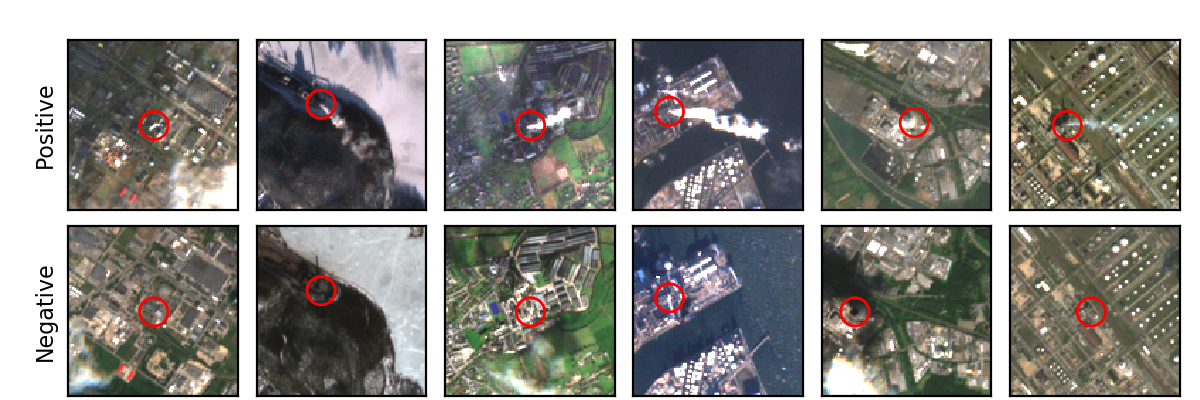}
  \caption{{\small A representative sample of example images. 
  Each column corresponds to a different location; the top row shows locations when a smoke
  plume is present (positive class), 
  the bottom row shows the same locations during the absence of smoke (negative class). 
  Red circles indicates the approximate origin of the plume. Our data set samples a wide range of seasonal effects, climate zones, land use types, and natural cloud patterns.}
  \label{fig:example_panel}}
  \vskip -0.4cm
\end{figure}

%
%
\section{Industrial Site Data Set}
\label{sec:data}

We acquire geographic locations of 624 sites from the European Pollutant Release and Transfer Register \citep{EPRTR2020}, a pollution reporting entity within the European Union as the foundation of our work.
For each site, we retrieve all available Sentinel-2 satellite imagery \citep{MSIuserguide} taken during 2019. Each raster image consists of all 12 spectral-band channels from the calibrated
(bottom-of-atmosphere) Level-2A reflectances and is cropped to an edge length of 1,200~m, corresponding to 120$\times$120 pixels. 
Low-resolution channels are resampled to the highest available resolution (10~m/pixel). A representative set of example images is shown in Figure \ref{fig:example_panel}.

Each image is manually classified based on the presence or absence of smoke plumes anywhere in that image; we further produce segmentation masks for 
1,437 random positive images with smoke plumes. 
We define as a smoke plume any opaque or semi-transparent cloud that originates from the surface; we make no assumptions on the plume's properties (e.g., its molecular composition or process of origin) other than it being most likely anthrophogenic.
Images (positive and negative) can include partial cloud coverage, however, images that are fully covered by natural clouds are excluded from the data set. We note that despite great care, the annotation process is highly subjective due to the low image resolution (10~m), scene variability caused by weather and illumination differences, and the fact that human annotation is only based on RGB images and does not consider additional spectral information present in the data. 
As a result, surface features such as buildings, ice, snow or partially occlusion by natural clouds might confuse a human annotator during the labeling process.

The final data set includes 21,350 images for 624 different locations with 
3,750
positive (a smoke plume is present) and 17,600 negative (no smoke plume is visible). For each location at least 1 image (negative only) and up to 96 images (positives and negatives combined) from the time-span during 2019 are available. We split the data into static subsets for
training (70\%), validation (15\%), and testing (15\%) in such a way that data for each location appear only in one of the three subsets.

%
%
\section{Classification: Identification of Smoke Plumes}
\label{sec:classification}
We investigate whether it is possible to reliably detect smoke plumes in our data given the challenges of natural clouds, ice and snow reflections, utilizing a ResNet-50 \citep{He2016} architecture as a binary classifier. The architecture is modified to utilize a 12 multi-band channel (all Sentinel-2 channels) input vector and results in a scalar logit. 
We use a binary cross-entropy loss function, which is minimized using stochastic gradient descent with momentum. 
The training and validation samples are balanced through duplication of the positive samples. Data augmentation is implemented in the form of random image mirroring and flipping, random
image rotations $i\cdot 90^\circ, i \in \{0, 1, 2, 3\}$, and the random cropping of a 90$\times$90~pixel window from each image.

After successful training from scratch we can report an accuracy of 94.3\% on the test data.
The confusion matrix is mostly symmetric with typical [TP/TN/FP/FN] ratios of [46.7\%, 47.6\%, 2.4\%, 3.3\%] underlining the reliable detection of smoke plumes from the Sentinel-2 satellite data.

\begin{figure}[t]
  \centering
  \includegraphics[height=1.9in]{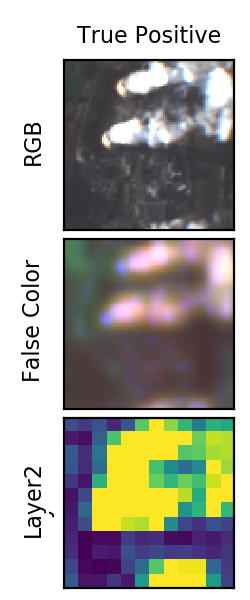}
  \includegraphics[height=1.9in]{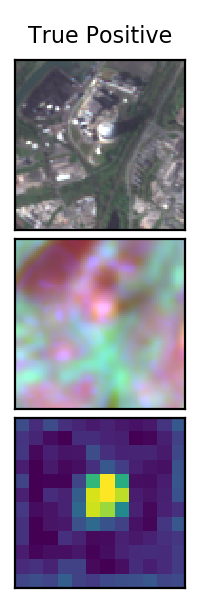}
  \includegraphics[height=1.9in]{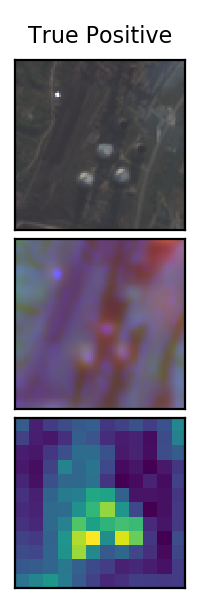}
  \includegraphics[height=1.9in]{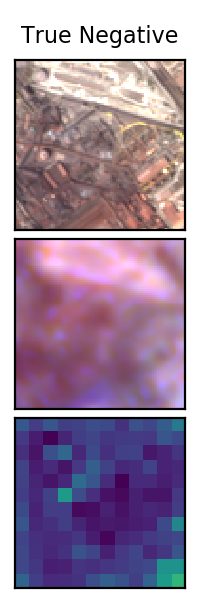}
  \includegraphics[height=1.9in]{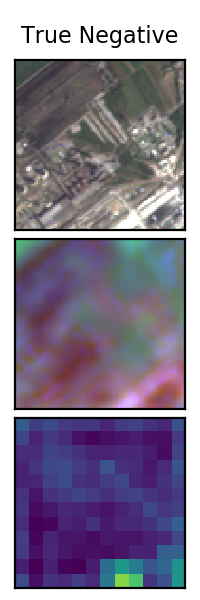}
  \includegraphics[height=1.9in]{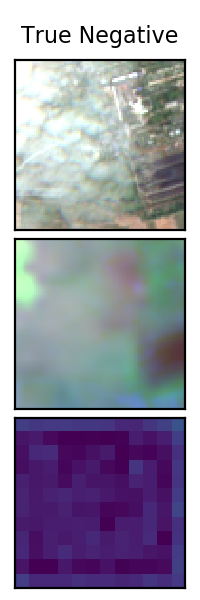}
  \includegraphics[height=1.9in]{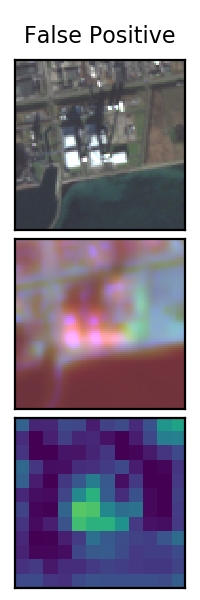}
  \includegraphics[height=1.9in]{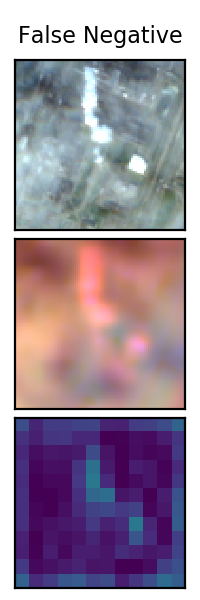}     
  \caption{{\small Evaluation of our classification model. For different examples from our 
  test sub-sample (columns), we show the true color RGB image (top row), a false color image 
  (center row, see Section \ref{sec:classification} for details), and the activations of Layer2 in our ResNet implementation
  (bottom row, sharing the same scaling across the row). We find that the location of smoke 
  correlates in most cases with significant aerosol and water vapor signals
  and that Layer2 activations fire based on these signals, leading to good localization of the smoke. 
\label{fig:class_results}}}
\vskip -0.4cm
\end{figure}

To investigate the decision-making process learned by our model, we sum up the gradients in the model's first convolutional layer (input layer) on a per-input-feature basis and find that in
most cases the mean gradients are highest for channels 1 (aerosols), 9 (water vapor), and 11 (Short-Wave Infrared band 1), which are reasonable since smoke plumes do release small particles and water vapor from the burning of fossil fuels \citep[see, e.g.,][]{Artanto2012}.
In Figure \ref{fig:class_results}, we find that the distribution of signal strength in these channels (center row, false color palette: red = channel 1, green = channel 9, blue = channel 11) correlates well with the locations of smoke plumes. 
Our model has learned to distinguish between smoke plumes and natural clouds (see 6th column in Figure \ref{fig:class_results}). 
The presence of high-altitude cirrus clouds seems to impede the 
detection of smoke plumes (see last column in in Figure 
\ref{fig:class_results}).
Finally, we find that activations in \texttt{Layer2} (the second bottleneck block in our ResNet-50) correlate well with the locations of smoke plumes in the image data (see Figure \ref{fig:class_results}, bottom row).

\begin{figure}[t]
  \centering
  \includegraphics[height=1.9in]{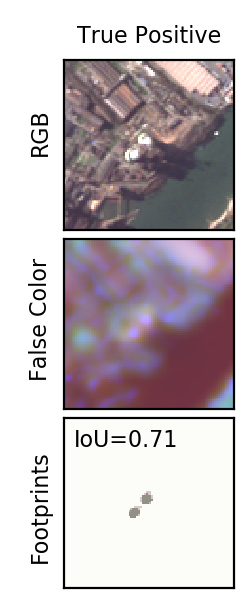}
  \includegraphics[height=1.9in]{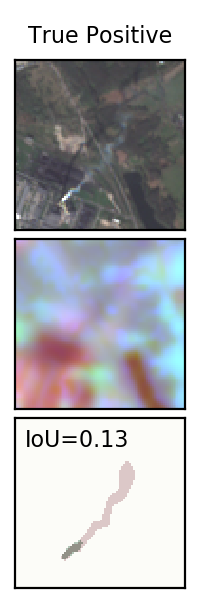}
  \includegraphics[height=1.9in]{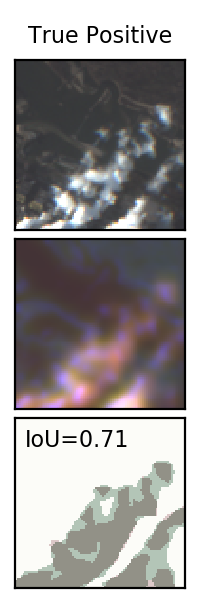}
  \includegraphics[height=1.9in]{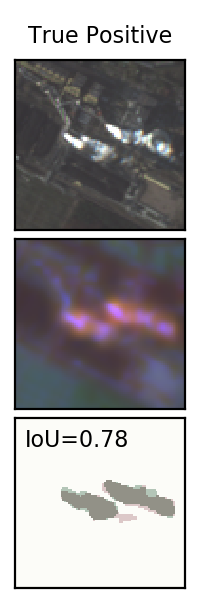}
  \includegraphics[height=1.9in]{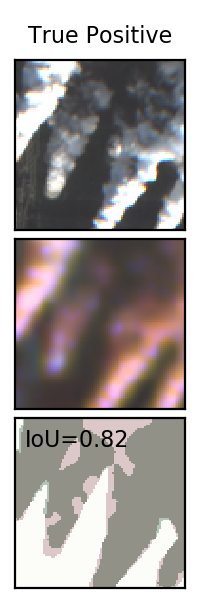}
  \includegraphics[height=1.9in]{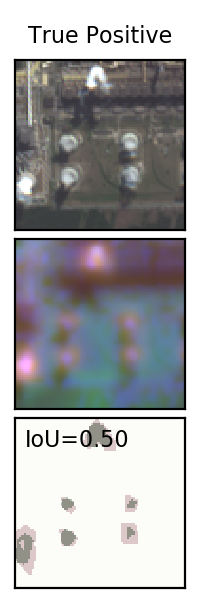}
  \includegraphics[height=1.9in]{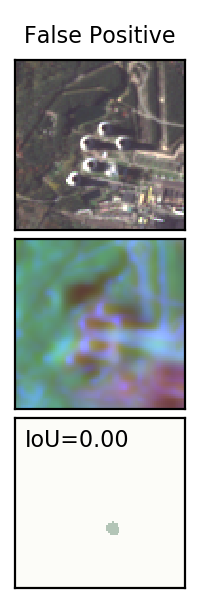}
  \includegraphics[height=1.9in]{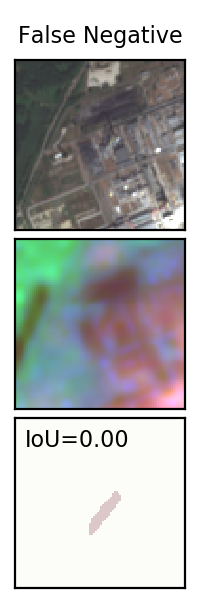}      
  \caption{{\small Evaluation of our segmentation model. For different examples from our 
  test sub-sample (columns), we show the RGB image (top row), a false color image 
  (center row, see Section \ref{sec:classification} for details), and the footprint of the ground-truth labels (red areas) 
  and predicted labels (green areas). While opaque smoke plumes are easily identified by the 
  model (columns 1, 3, 4, and 5), semi-transparent smoke is sometimes missed by the model 
  (columns 2 and 6). We also find that 
  false negatives (column 8) are often related to high-altitude cirrus clouds and that false positives (column 7)
  are mainly caused by human mislabeling (see Section \ref{sec:data}).
\label{fig:seg_results}}}
\vskip -0.4cm
\end{figure}

%
%
\section{Semantic Segmentation: Quantifying Smoke Plumes}
\label{sec:segmentation}

We investigate whether it is possible to quantify the amount of smoke present in an image through semantic segmentation; the measurement of the number of pixels occupied by smoke may be used as a proxy for the level of industrial activity in this area.
We utilize a U-Net \citep{Ronneberger2015} implementation in combination with a binary cross entropy loss function and stochastic gradient descent with momentum to learn the segmentation task.
This model is learned on a training data set comprising 70\% of our sample of 
1,437 images with manually labeled smoke plumes (see Section \ref{sec:data}) to which we add the same number of negative images (no smoke plumes present). The remaining labeled images are evenly split into a validation subset and a test subset to which we add negative examples in the same way.
The performance is evaluated using both the accuracy metric, in which we consider images that 
contain any amount of smoke positive or negative otherwise, and using the Intersection-over-Union (IoU, also known as Jaccard index) metric, in which we ignore images that do not contain smoke labels as those have an ill-defined IoU metric.  
Our trained model achieves an accuracy 94.0\% and an IoU of 0.608 based on our test sub-sample. Finally, we find that on average 94.4\% of the area covered by smoke in each positive image is reproduced by our model.

Figure \ref{fig:seg_results} shows example images after running through our segmentation model.  We find that the model reliably finds opaque smoke plumes while it is less reliable in finding somewhat transparent smoke plumes; occasionally, surface objects are mistaken as smoke plumes. 
We find a saturation training sub-sample IoU of ${\sim}$0.7, which we attribute to the model's issues with semi-transparent clouds, as well as short-comings in the manual annotation of the data (see Section \ref{sec:data}). 

We conclude that using our approach we can  measure the extent of smoke plumes (on average to within 5.6\% of human performance), which is a prerequisite for estimating actual emission levels from our results. This will be further investigated in future work in the form of an analysis involving smoke plume area estimates from our segmentation model and ground-truth activity metrics from select power plants and other emitters. This process will require to take into account environmental factors \citep[e.g., as provided by][]{ERA5}.



\end{document}